\documentclass{article}
\usepackage[preprint]{icml2025}    % arixv version

\usepackage[inline]{trackchanges}

\usepackage{wrapfig}
\usepackage{comment}
\usepackage{colortbl}

\usepackage{hyperref}

\usepackage{amsthm} % unnumbered theorem etc
\usepackage{amsmath} % WC: aligned equations
\usepackage{color,soul}
\usepackage{graphics}
\usepackage{lipsum}
\usepackage{tikz}

\usepackage{enumitem}	% for fine control of list env
\usepackage{listings} % for code listing
\usepackage{booktabs}	% xzl: for fancy tables
\usepackage{lipsum}

\usepackage{capt-of}	% xzl: used to override the ``figure'' and ``table'' in caption.
\usepackage{diagbox}
\usepackage{multirow}
\usepackage{todonotes}  %xzl: for "missing figures"
\usepackage{subcaption} %xzl: for subfigure. incompatible with package subfig

\definecolor{refkey}{rgb}{249,158,26}
\definecolor{labelkey}{rgb}{0,1,0}
\definecolor{airforceblue}{rgb}{0.36, 0.54, 0.66}
\definecolor{applegreen}{rgb}{0.55, 0.71, 0.0}
\definecolor{celestialblue}{rgb}{0.29, 0.59, 0.82}
\definecolor{navyblue}{rgb}{0.29, 0.59, 0.82}
\definecolor{cerulean}{rgb}{0.0, 0.48, 0.65}
\definecolor{darkorchid}{rgb}{0.6, 0.2, 0.8}
\definecolor{electriccyan}{rgb}{0.0, 1.0, 1.0}
\definecolor{forestgreen(web)}{rgb}{0.13, 0.55, 0.13}
\definecolor{labelkey}{rgb}{0,0,1}
\definecolor{refkey}{rgb}{0,0,1}

\usepackage{mathrsfs}	% xzl: for Ralph Smith’s Formal Script Font (rsfs)
\usepackage{amsfonts}

\usepackage[export]{adjustbox} % xzl: flush figure to left/right

\usepackage{boldline}					% jin: use boldline in the table

\usepackage{multirow}

\usepackage[capitalize,noabbrev]{cleveref}

\theoremstyle{plain}

\theoremstyle{definition}

\theoremstyle{remark}

\definecolor{frenzyorange}{RGB}{249, 158, 26}

\newcommand{\fcircle}[2][red,fill=red]{\tikz[baseline=-0.5ex]\draw[#1,radius=#2] (0,0) circle ;}%
\newcommand{\fdiamond}[2][red,fill=red]{%
  \tikz[baseline=-0.5ex]\draw[#1] (0,0) -- (#2,#2) -- (0,2*#2) -- (-#2,#2) -- cycle;%
}
\newcommand{\fsquare}[2][red,fill=red]{%
  \tikz[baseline=-0.5ex]\draw[#1] (0,0) rectangle (#2,#2);%
}
\newcommand{\ftriangle}[2][red,fill=red]{%
  \tikz[baseline=-0.5ex]\draw[#1] (0,0) -- (#2,0) -- (#2/2,{sqrt(3)*#2/2}) -- cycle;%
}

\renewcommand{\paragraph}[1]{\vskip 3pt\noindent\textbf{#1 }}	 % used to be 6pt

  {\begin{list}{$\bullet$}%
     {\setlength{\parsep}{0pt}%
      \setlength{\topsep}{0pt}%
      \setlength{\itemsep}{2pt}}}%
  {\end{list}}
\newcommand\Noted[1]{} % remove highlights.

\definecolor{darkblue}{rgb}{0.0, 0.0, 0.55}
\definecolor{mygreen}{HTML}{ADFF2F}
\definecolor{mylightgray}{gray}{0.8}

\newenvironment{myitemize}%
  {\begin{itemize}
	[leftmargin=0cm,
		itemindent=.3cm,
		labelwidth=\itemindent,
		labelsep=0pt,
		parsep=1pt,
		topsep=1pt,
		itemsep=1pt,
		align=left]
  }%
  {\end{itemize}}    

  {\begin{enumerate}
	[leftmargin=.cm,itemindent=.5cm,labelwidth=\itemindent,
		labelsep=0pt,
		parsep=1pt,
		topsep=1pt,
		itemsep=3pt,
		align=left]
  }%
  {\end{enumerate}}    

\newcommand{\sys}{RWKV-Lite}
\begin{document}
% ****************** TITLE *******************************************
\twocolumn[
\icmltitle{\sys{}: Deeply Compressed RWKV for Resource-Constrained Devices}

% ****************** AUTHORS **************************************


\icmlsetsymbol{equal}{*}

\begin{icmlauthorlist}
\icmlauthor{Wonkyo Choe}{uva}
\icmlauthor{Yangfeng Ji}{uva}
\icmlauthor{Felix Xiaozhu Lin}{uva}
\end{icmlauthorlist}

\icmlaffiliation{uva}{Department of Computer Science, University of Virginia, Charlottesville VA, USA}

\icmlcorrespondingauthor{Wonkyo Choe}{wonkyochoe@virginia.edu}
%\icmlcorrespondingauthor{Firstname2 Lastname2}{first2.last2@www.uk}

\icmlkeywords{LLM, Compression}

\vskip 0.3in
]

\printAffiliationsAndNotice{}

% ----------------------------------------------

%\pagestyle{plain} % should come right after \maketitle
% !TeX root = main.tex

\begin{abstract}

To deploy LLMs on resource-contained platforms such as mobile robots and smartphones, non-transformers LLMs have achieved major breakthroughs.
Recently, a novel RNN-based LLM family, Repentance Weighted Key Value (RWKV)~\cite{peng2023rwkvreinventing,peng2024eaglefinch} 
has shown strong computational efficiency; 
nevertheless, RWKV models still have high parameter counts which limited their deployment.
In this paper, we propose a suite of compression techniques, 
ranging from model architecture optimizations to post-training compression, tailored to the RWKV architecture.
Combined, our techniques reduce the memory footprint of RWKV models by 
3.4x -- 5x with only negligible degradation in accuracy; 
compared to transformer LLMs with similar accuracy, our models require 4x less memory footprint.

\end{abstract}

% !TeX root = main.tex

\section{Introduction}
\label{sec:intro}

% !TeX root = main.tex
\begin{figure}
	\centering
	\includegraphics[width=0.45\textwidth]{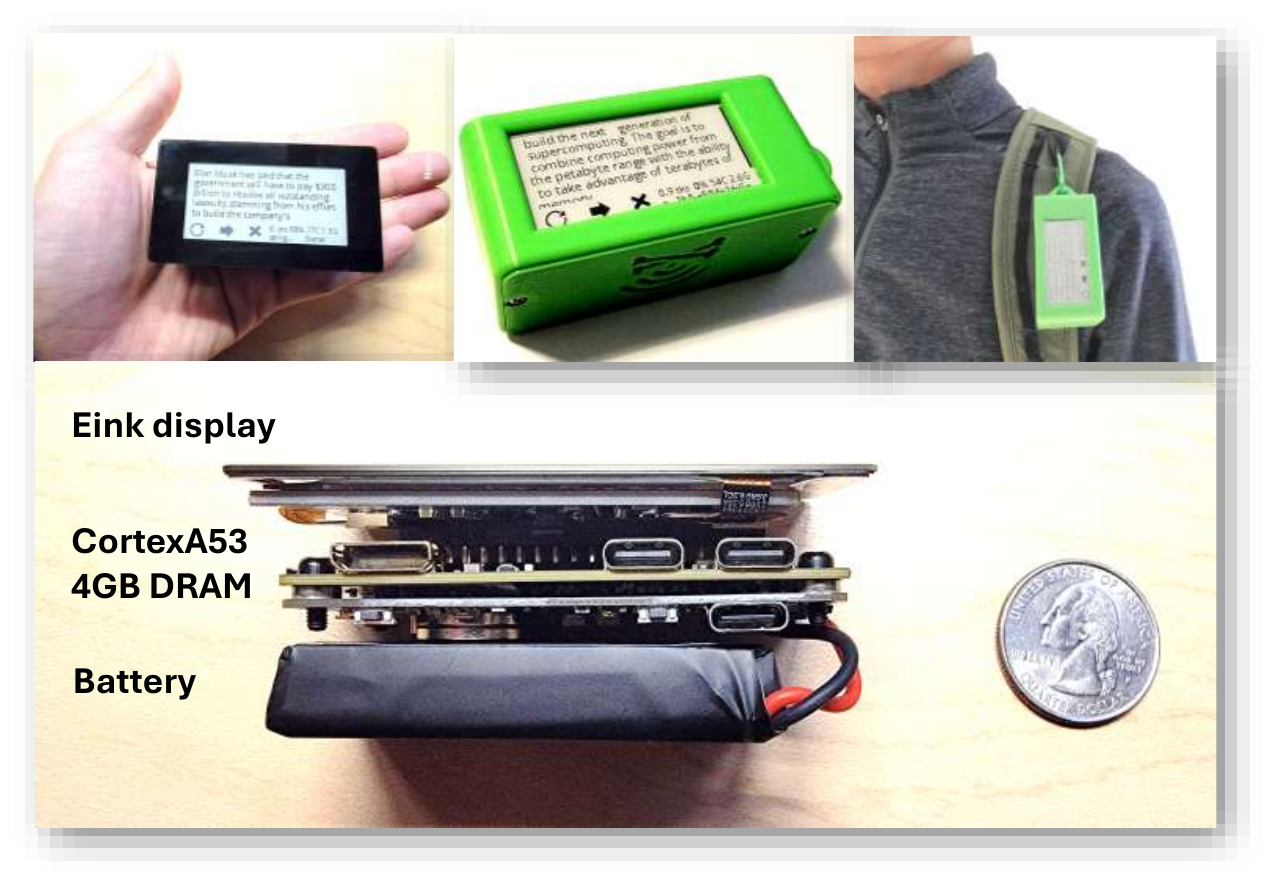}
    \caption{A proof-of-concept system we built, 
    which runs the compressed RWKV model reported in the paper 
    and demonstrates the concept of running LLMs on wearable devices.
    See Table~\ref{tab:platform} for hardware details.
    }
	\label{fig:demo}
\end{figure}
While the transformers shown superior performance in 
large language models (LLMs),
they are computationally expensive and often require a large amount of memory.
Even smaller models, such as Llama 2-7B with INT4, require about 5GB of memory and 0.11 tokens/sec on Raspberry Pi 4~\cite{dhar2024empiricalanalysis}.
Thus they are beyond the capacity of many edge devices 
such as wearable gadgets, drones/humanoids, and smartphones/tablets. 
These devices have a strong need for LLMs, while invoking the cloud LLMs is often undesirable.

To this end, one compelling alternative to transformers is an RNN-based LLM family, RWKV~\cite{peng2023rwkvreinventing,peng2024eaglefinch}.
Beyond classic RNNs for language modeling~\cite{sherstinsky2020fundamentalsrecurrent,mikolov2010recurrentneural,lstm},
RWKV incorporates \textit{multi-headed vector-valued states} and dynamic recurrence mechanisms. 
It achieves accuracy comparable transformer-based LLMs while still retaining high inference efficiency as in RNNs. 
For instance, on ARM Cortex-A76  processors,
RWKV 7B was reported to generate 16.39 tokens per second~\cite{githubGitHubRWKVrwkvcpp}, 
whereas Llama-7B only generates several tokens per second. 
It has been reported that RWKVs are already pre-installed on millions of Windows 11 machines~\cite{rwkv_windows}.

Despite their potential for edge devices, 
RWKV models are as parameter-heavy as transformer models (see Section \ref{sec:motiv} for an analysis).
During inference, the RWKV 1.5B model requires about 4GB of memory and a few GBs of memory, even after quantization.
This high memory footprint makes RWKV less practical for wearables and entry-level mobile devices, creating a major deployment barrier.

\textbf{The goal of this paper} 
is to reduce the memory footprint of RWKV models during inference. 
To achieve this, we propose the model architectural optimizations and recover accuracy through continual learning, 
as well as post-training techniques that dynamically load only a small subset of the model parameters.
Specially, our techiques are: 
\begin{myitemize}
    \item For projection weight matrices in the RWKV blocks (channel-mix or time-mix), we apply low-rank approximations catering to either continual training or pre-training. 
    \item For FFNs in the channel-mix layers, we identify the existence of sparsity and propose a novel sparsity predictor,
    which ensembles a classic MLP (multi-layer perceptron) 
    and a 1-bit quantized FFN. 
    \item To compress the embedding and classification head layers, we propose embedding cache and hiereachical weight decomposition, respectively; they particuarlly benefit smaller RWKV models.
\end{myitemize}
Altogether, our techniques reduce the memory needed by RWKV inference by 3.4x -- 5x, while only seeing negligible degradation in accuracy and perplexity.
Combined with model quantization, our techniques achieve an end-to-end memory reduction of up to 10x, compared to the vanilla models.
Our optimizations, which focus on fundamental RWKV building blocks, are applicable to the latest RWKV version 7 that emerged concurrently with our work~\cite{coda-fornometaincontextlearning}.
% !TeX root = main.tex

\section{Motivations}
\label{sec:motiv}

% !TeX root = main.tex
\begin{figure}
	\centering
	\includegraphics[width=0.45\textwidth]{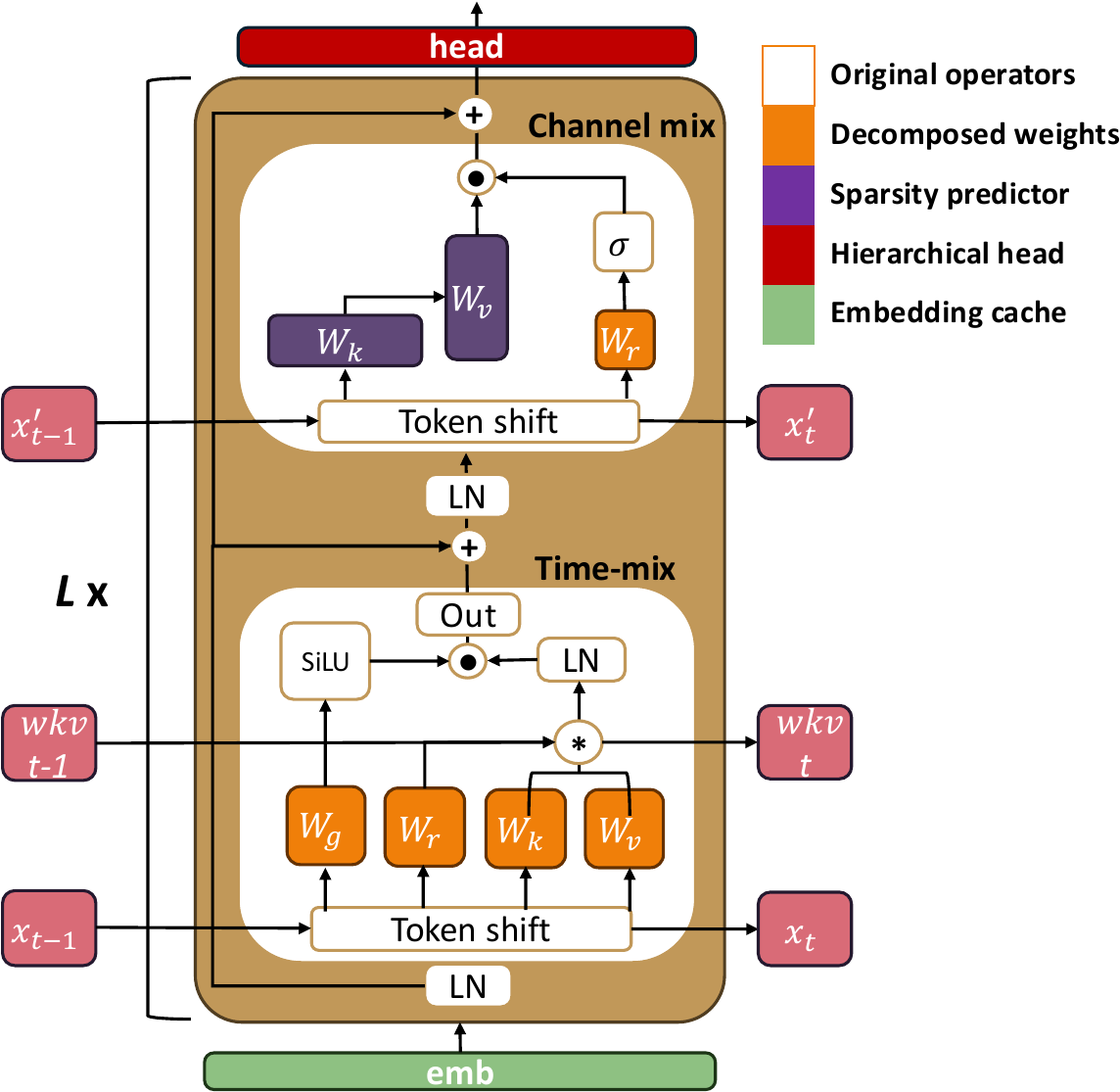}
    \caption{
    Simplified architecture of the RWKV model. 
    Each variant has multiple ($L$) numbers of RWKV blocks, which comprise time and channel-mix layers.
    Colored blocks are our techniques onto the original layers.
    (LN=Layer Normalization).
    }
	\label{fig:rwkv}
\end{figure}

\begin{table}[h]
	\centering
    \caption{The parameter distribution of RWKV models, 
    focusing on the 0.1B--3B variants suitable to edge devices. 
    Square matrices are defined as $W \in \mathbb{R}^{D \times D}$, and non-squares are $W \in \mathbb{R}^{D \times 3.5D}$.
    }
	\includegraphics[width=0.48\textwidth]{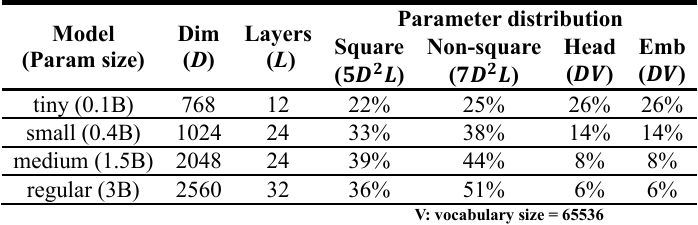}
	\label{tab:param-dist}
    \vspace{-0.4in}
\end{table}

\subsection{RWKV architecture}
Just like transformer-based LLMs, 
each RWKV model includes an embedding layer at the bottom (near the input), 
and a classification head at the top (near the output).
Unlike transformers, 
between the embedding and the header are 
a stack of RWKV blocks (e.g. 12 or 24), each of which consists of a channel-mixing and a time-mixing layer (which are often referred to as ``Attention (attn)'' and ``Feed-Forward Network (FFN)'' for convenience). 
As depicted in Figure~\ref{fig:rwkv}: 
RWKV blocks eschew attention mechanisms; 
rather, the state information propagates across timesteps as small, fixed-size vectors (e.g. $wkv_{t-1}$ or $wkv_t$ elements). 
Specifically, channel-mixing layers take the role of short-term memory to store the state of the previous information, while time-mixing layers act as long-term memory to retain a part of the previous states over a longer period.
Inside RWKV blocks, the majority of weights are in the projection matrices termed as Receptance, Weight, Key, and Value, which are analogous to (but functionally different from) the Query, Key, and Value weights of Transformers.

\subsection{Memory footprint analysis}

\paragraph{Overall parameter distribution}
RWKV models, depending on their variants, have different weight distribution across layers. 
Overall, the size of RWKV blocks scale quadractically with dimension $D$ and the number of blocks $L$.
Depending on the block configuration, its weight matrix has a size of either $W \in \mathbb{R}^{D \times D}$ (square) or $W \in \mathbb{R}^{D \times 3.5D}$ (non-square). 
By comparison, 
Each of the embedding and the head layer is a $M \in \mathbb{R}^{D \times V}$ matrix; where 
$V$ is the vocabulary size. 
Embedding layer is a lookup table given a token index, and head layer is a linear layer to project the hidden state to a distribution over the vocabulary.
Their sizes only scale with $D$ but not with number of RWKV blocks.
As denoted in Table~\ref{tab:param-dist}, 
for smaller models (tiny and small), the embedding and head layers occupy almost a half or third of the parameter sizes; 
such a fraction diminishes to about 15\% for the medium model, 
for which the RWKV blocks dominate. 

This suggests that: 
for smaller models, compressing the embedding and head layers would be as important as compressing the RWKV blocks;
for larger models, however, compressing embedding and head will yield marginal benefit. 
This observation guides us for navigating through tradeoffs between model size for accuracy, as Section~\ref{eval:ablation}.

% !TeX root = main.tex
\begin{figure}
    \centering
    \includegraphics[width=0.4\textwidth]{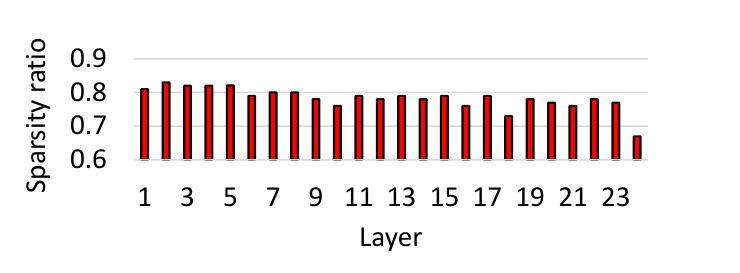}
    \caption{Average FFN's sparsity ratio (the fraction of zero values in activations),
    showing substantial sparsity across layers and unused weight row/columns were loaded to memory.
    Tested on 200 token generations in the channel-mix layer of the small RWKV model.}
    \label{fig:sparsity}
\end{figure}

\paragraph{Parameters in RWKV blocks}
The RWKV weights are dominated by two groups of weight matrices: 
(1) several square matrices $W \in \mathbb{R}^{D \times D}$ in channel mix and time mix; 
they constitute 22\% -- 39\% of total model weights, across different model sizes; 
(2) two FFN matrices for FFN $W \in \mathbb{R}^{D \times 3.5D}$ in channel mix;
they constitute 25\% -- 51\% of total model weights. 
Both groups of weights are critical for the model performance, and they need to be compressed. 
The sizes of all other weights (small vectors) are negligible.
This is summarized in Table~\ref{tab:param-dist}.

To compress the square projection matrices, 
our approach is low-rank approximation. 
Our hypothesis is that the intrinsic rank of such projection is low, and can be approximated by the product of two much smaller matrices (one 'tall' and one 'flat'). 
Such low-rank approximation has shown high efficacy in model compression (e.g. LoRA~\cite{hu2021loralowrankb} for fine-tuning LLMs); 
however, it has been unclear whether they would apply to RWKV, which has very different representation than transformers; 
it was also unclear what mechanism (and effort) would be needed to recover the accuracy lost due to decomposition.

To compress the FFN non-square matrices, our approach is to exploit sparsity. 
Note that the FFN matrices cannot be effectively compressed using low-rank approximation because they are not square and already have a relatively low rank. Further decomposition does not significantly reduce their size and quickly degrades accuracy.
Our motivation for exploiting sparsity is the nonlinearity (squared ReLU) between the two FFN matrices, which suppresses negative neurons as zeros,
as well as the lottery ticket hypothesis~\cite{frankle2019lotteryticketa} stating that 
only a small subset of parameters is essential for each token generation.
However, the sparsity of RWKV models has not been examined before, and the lottery ticket hypothesis has not been tested on RWKV, especially for smaller models~\cite{liu2023dejavu,xue2024powerinfer2fast,song2024turbosparsea,song2024powerinferfast,song2024prosparseintroducing}.

Our new findings validate the existence of sparsity in RWKV: 
to generate a token, 
only a small fraction of neurons in the FFN activation vector have non-zero values;
correspondingly, only a small fraction of rows/columns in the FFN weight matrices participate in the computation effectively, and only these rows/columns need to be loaded in memory for this token. 
Figure~\ref{fig:sparsity} shows the sparsity of the RWKV's FFN weight matrices across layers, 
ranging from 83\% (bottom layers) to 67\% (top layers).

To exploit the sparsity for memory saving, however, requires us to  
predict activated neurons without actually computing them. 
This raises significant, new challenges. 
(1) The RWKV sparsity level (--83\%) is notably lower than large transformers (e.g. reported to be 99\% in OPT-175B), which most prior work focused on \cite{liu2023dejavu}.
This not only means the "headroom" for memory reduction is lower than large transformers, 
but also means that finding actually activated neurons is more difficult. 
False positives in predicting activated neurons would result in loading unnecessary weights, quickly diminishing the saving; 
false negatives would miss actually activated neurons that are key to model inference -- leading to accuracy collapse. 
(2) As we focus on smaller models, 
the memory overhead of predictors themselves (often smaller neural networks) becomes non-negligible; 
the overhead of predictors could totally diminish the benefit of sparsity.
These challenges combined, sparsity predictors known effective for transformers will fail for RWKV, 
as we will show in Section~\ref{design:sparsity}.
% !TeX root = main.tex
\section{\sys{}: Deeply compressed RWKV}
\label{sec:design}
As shown in Figure~\ref{fig:rwkv}, 
\sys{} incorproates multiple techniques for memory footprint reduction. 

\subsection{Singular Vector Decomposition for RWKV blocks}
\label{sec:svd}
\label{sec:design:svd}

We transform the projection weight matrices in RWKV blocks:
$W_{r,k,v,g}$ in a time-mix layer, and $W_r$ in a channel-mix layer, as depicted in Figure~\ref{fig:rwkv}.
We do not transform $W_o$ in the time-mix, as doing so is detrimental to model performance.

Given a projection matrix $W \in \mathbb{R}^{M \times M}$, where $M$ is the embedding size,
we propose two methods motivated by SVD (equation \ref{eq:svd} in Appendix), 
suiting different use cases:

(1) Simple SVD with continual pretraining: given input data $X \in \mathbb{R}^{1 \times M}$, the projection with $W$ is represented as follows:
\begin{equation}
\label{eq:continue-pretrain}
XW \approx (XL)R
\end{equation}
in which: $L \in \mathbb{R}^{M \times (M/k)}$ is a weight matrix $U\Sigma$, having $M^2/k$ parameters; 
$R \in \mathbb{R}^{(M/k) \times M}$ is a weight matrix $V$, having also $M^2/k$ parameters, 
and $k$ is a compression factor, e.g., $k = 8$. 

To apply this approximation, we start from a vanilla pretrained model.
(i) We replace a linear projection $W$ with two smaller linear layers: $L$ and $R$, for which the initival values are from solving SVD of $W$ with retaining only the top $M/k$ singular values.
The approximation reduces the projection's parameters from $M^2$ to $2M^2/k$. 
(ii) We continue to pre-train the model with the decomposed weights.
Such continual training (not task-specific fine-tuning) is vital to recover accuracy, 
because the decomposed weights have lower ranks than $W$ and therefore smaller capacity.

(2) Enhanced SVD with normal pretraining: 
to overcome the limited capacity of $L$ and $R$, 
the projection with $W$ can be replaced as below:
\begin{equation}
\label{eq:regular-pretrain}
XW \approx \text{ReLu}(XL)^2 R + (XD)
\end{equation}
with $D$ as a full-rank, diagonal matrix. 

We hypothesize that such a new construct would improve the model expressiveness for approximiting the original $W$, while still significantly shrinks the parameters (roughly by $k\times$).
Specifically: squared ReLu's non-linearity amplifies significant activations and suppressing irrelevant ones; 
the diagonal matrix compensates for the lower ranks of $L$ and $R$.
On the flip side, the values of $L$, $R$, and $D$ can no longer be conveniently derived from a pretrained $W$ by solving its SVD.  
We therefore use this architecture only for pretraining RWKV from scratch. 

\paragraph{Applying the techinques}
After continual training or pretraining, 
$L$, $R$, and (optionally) $D$ are saved to the model file, while $W$ is discarded; 
at inference time, $L$, $R$ and $D$ are loaded to memory for projecting input vector $X$, as in Equations ~\ref{eq:continue-pretrain} and ~\ref{eq:regular-pretrain}.

\paragraph{Computational overhead consideration}
Compared to the vanilla RWKV, 
our two methods indeed require additional matrix-vector multiplications 
and/or element-wise operations (via ReLU and the diagonal matrix operation). 
Yet, the overhead is either low or overshadowed by other operations in the model. 
In the end, the added wall time is small, as shown in Appendix, Figure~\ref{fig:infer-inhouse}.

\subsection{Leveraging sparsity for FFNs in channel-mix}
\label{design:sparsity}

For FFNs in channel-mix (i.e. non-square matrices $W_{k,v}$), 
we utilize sparsity to load much fewer weights at inference time.
Our approach hinges on predicting neurons 
(defined as columns in $W_k$ and rows in $W_v$)
needed for a given input token. 

Given an input $X \in \mathbb{R}^{1 \times M}$, 
we start by using a trainable MLP predictor $P_{MLP}$, 
as commonly used for predicting transformer sparsity~\cite{xue2024powerinfer2fast, liu2023dejavu}. 
The sparsified FFN computation are as follows: 
\begin{equation}
    \begin{split}
        P_{MLP} = 1_{\sigma(\text{relu}(XL_1)L_2)\ge t}, \\
        \text{FFN} (X) \approx \text{relu}(XW_k \cdot P_{MLP} )^2 \cdot W_v \\
    \end{split}
\end{equation}
where $L_1 \in \mathbb{R}^{M \times N}$, and $L_2 \in \mathbb{R}^{N \times M}$ are linear projections trained from scratch. $M$ is the embedding size, $N$ is a hidden dimension for a linear layer, and $\sigma$ represents the sigmoid activation function.
$P_{MLP}$ is a binary vector of size $M$ that predicts the neurons to be activated in $W_k$ where its elements are determined by the threshold $t$, e.g,. $P_{MLP, i} = 1$ if $\sigma(\text{relu}(XL_1)L_2) \ge t$; otherwise, 0.
Note that since the computation for $W_k$ is sparsified, the second computation for $W_v$ is sparsified as well.

While these trainable MLP predictors performed well on transformer FFNs, 
we found their efficacy is limited on RWKV, resulting in either low recall 
(missing too many activated neurons; bad model accuracy) or low precision (too many inactivated neurons loaded; marginal memory saving). We hypothesize the causes as: RWKV's lower sparsity than transformers (Figure~\ref{fig:sparsity}); RWKV's unique embedding space; that the RWKV models under question are already small which limit the room to save. 

To this end, we complement an MLP predictor with a deeply quantized version of the FFN itself. 
Given an input $X \in \mathbb{R}^{1 \times M}$, the quantization-based predictor is: 
\begin{equation}
    P_{quant_i} = 1_{XW^{\text{INT(i)}} \ge t}, \quad i \in \{1, 2, ...\}, 
\end{equation}
where $W^{INT(i)}$ is the quantized weight matrix with $i$ bits, and threshold $t$ is a percentile, e.g., $P_{quant1, i} = 1$ if $XW^{INT(1)}$ is greater than or equal to the 80th percentile (t=0.8) of $XW^{\text{INT}(1)}$.
In practice, we find $i=1$ is sufficient, which makes $P_{quant_i}$ one order of magnitude smaller than the FFN itself. 

The final prediction ensembles MLP and 1-bit quantization: 
\begin{equation}
    \begin{split}
    P_{ens} = \text{max}(P_{MLP}, P_{quant_i}) \\
    \text{FFN} (X) \approx \text{relu}(XW_k \cdot P_{\text{ens}} )^2 \cdot W_v 
    \end{split}
\end{equation}
Interestingly, we find that using the 1-bit predictor alone still leads to poor accuracy. 
Upon closer inspection, we observed that MLPs can identify most activations with moderate values but may miss a small number of activations with high values. 
Conversely, the 1-bit predictor can reliably predict these high-value outliers but tends to err on neurons with moderate values near the decision boundary.

\paragraph{Applying the techinques}
Given a frozen RWKV model, 
one ensemble of $\langle P_{MLP}, P_{quant_i} \rangle $ is initialized for each FFN. 
They are trained with supervision from the RWKV model. 
(See Section~\ref{sec:impl} for training details).
At inference time, given an input $X \in \mathbb{R}^{1 \times M}$, 
an FFN runs its predictor ensemble, and only loads the predicted FFN neurons. 

\subsection{Embedding \& classification heads}
\label{design:outer}
To smaller models, the embedding and head layers are of particular interest 
because each contributes substantial parameters, e.g., 26\% and 14\% for 0.1B and 0.4B models respectively (Table~\ref{tab:param-dist}). 

\paragraph{Embedding caches}
We build a cache for embeddings of recent input tokens. 
The cache requires no training. 
When exceeding its capacity (in our implementation, 1000 embeddings which are only 1.5\% of the embedding layer), the cache evicts the least recently used entries. 
Caching works because token usage in language follows a long-tail distribution~\cite{jozefowicz2016exploringlimits}, where a small number of tokens frequently appear.
Similar embedding caches were explored before, e.g. in retrieval systems to accelerate retrieval processing and reduce the memory costs~\cite{jin2024ragcacheefficient}.
\paragraph{Hierearchical heads}
Compressing the head $H$ is more challenging: 
for each input token, 
the head uses all its weights to compute a probablity distribution over for all tokens in the vocabulary.
Our approach during inference is (1) to compute \textit{exact} logits for highly probable tokens,
for which we load the needed weights; 
and (2) to generate \textit{pseudo} logits for other tokens, without loading any weights.

% !TeX root = main.tex
\begin{figure}
	\centering
	\includegraphics[width=0.48\textwidth]{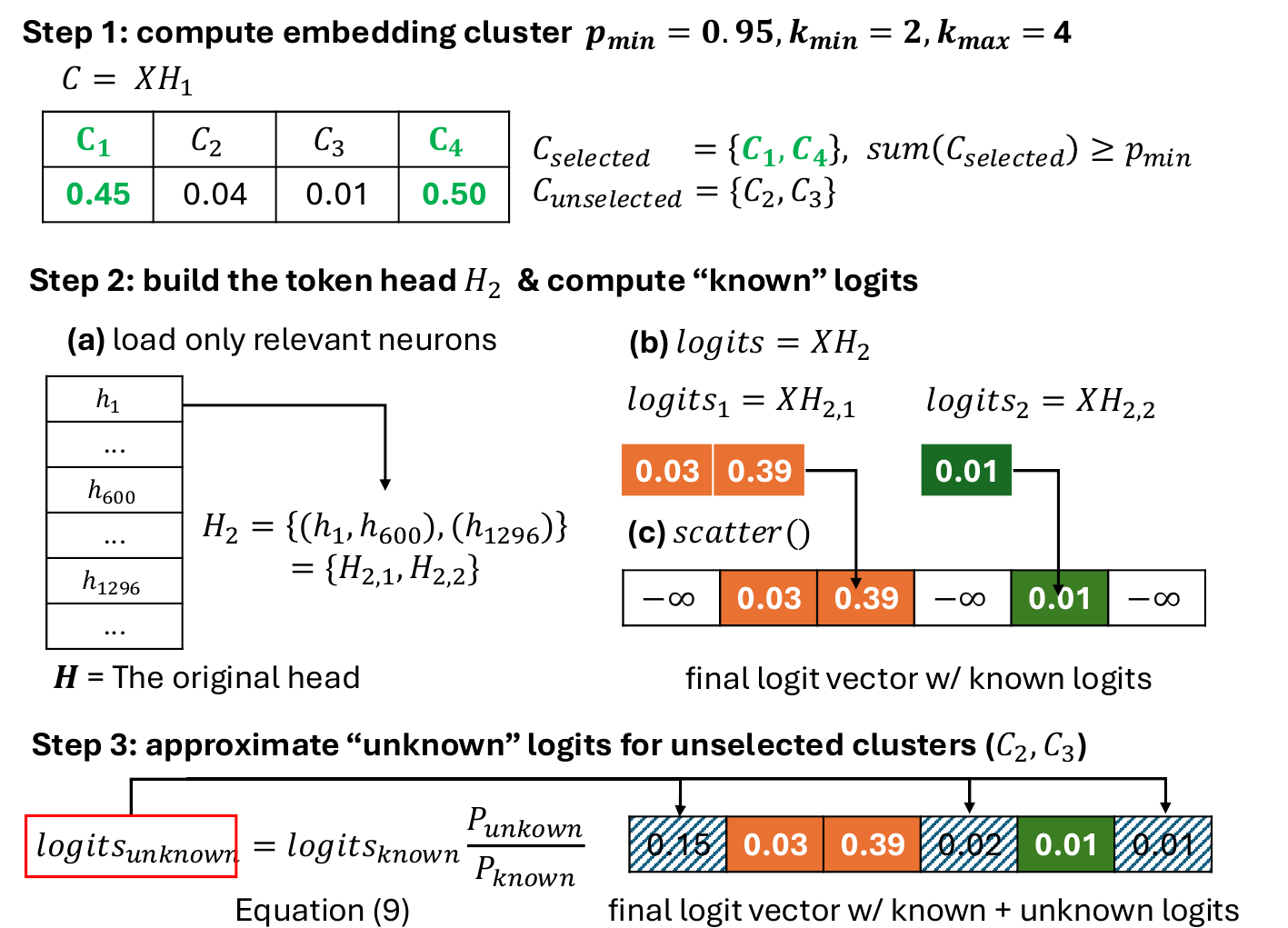}
	\caption{
    An illustration of hierarchical heads, comprising a cluster head and many per-cluster token heads. 
    At inference time, it computes the probability of each cluster, selects most probable clusters and selectively load their token heads to compute the token logits (orange/green boxes). Finally, it computes pseudo logits for tokens in unselected clusters (blue-pattern filled box).
    }
	\label{fig:clustered-head}
\end{figure}
Starting from a frozen RWKV model, 
we apply offline clustering of all the trained token embeddings, e.g. 64K in total. 
We use simple K-means clustering based on embedding Euclidean distances, yielding a set of $N$ clusters (e.g., $N$ set to 200 in our implementation). 
We then initialize two levels of new classification heads to be used in place of the original head during inference: 
(1) a cluster head $H_1 \in \mathbb{R}^{M \times N}$, which computes the \textit{aggregated} probability of each cluster given an input token, and 
(2) $N$ token heads, $H_2 = \{H_{2,1}, H_{2,2}, \dots, H_{2,N}\}$, where $H_{2,i} \in \mathbb{R}^{M \times T_i}$ 
and computes the logits of all $T_i$ tokens within the $i$-th cluster. 
We train the cluster head $H_1$ with supervision from the original head $H$, with loss defined as:
\begin{equation}
    D_{\text{KL}}\left(\bar{H} \parallel H_1\right) = \sum_{i=1}^{V} \bar{H} \log \frac {\bar{H}} {H_1},
\end{equation}
where $\bar{H}$ is a derived form of $H$ that each element is the sum of token probabilities in its cluster.
We use the KL-divergence loss to quantify difference of probability distribution between $H_1$ and $\bar{H}$; in our case, the vocabulary size $V$ is 65536.
The training is light, see Section~\ref{sec:impl} for details. 

We do not train the token heads $H_2$; instead, we directly copy their weights from the corresponding token rows of the original head $H$.
For instance, if one selected cluster contains two tokens, their original weights in $H$ are copied to $H_2$.

At inference time, given an input $X \in \mathbb{R}^{1 \times M}$, 
our hierarchical head computes in three steps
(as shown Figure~\ref{fig:clustered-head}):

Step 1: compute the probabilities over all $N$ clusters:
\begin{equation} \label{eq:cls-1} 
    C = \text{softmax}(XH_1)
\end{equation}
that is, the trained cluster head $H_1$ projects $X$ to the cluster probability space.

Our head then selects the most probable $k$ clusters from $C$.
The selection is made such that the cumulative sum of cluster probabilities exceeds $p_{min}$,  
and that the total number of selected clusters exceed $k_{min}$ but no more than $k_{max}$.
$p_{min}$, $k_{min}$, and $k_{max}$ are predefined thresholds, e.g., 0.95, 3, and 100 in our implementation.
Our rationale is that sampling tokens from sufficiently diverse clusters (as opposed to only from the most probable cluster) is crucial to not missing ``good'' tokens in generation. 

Step 2: for each selected cluster, compute the token logits:
\begin{equation} \label{eq:cls-2} 
    \text{logits}_i \approx X \cdot H_{2,i}  \quad \forall i \in \{1, 2, \dots, k\}
\end{equation} 
Based on the selected clusters, we extract relevant weights $H_2 = \{H_{2,1}, H_{2,2}, \dots H_{2,k}\}$. 
This step is how we save memory: 
it loads only the token heads for the selected clusters, which are much smaller than the original head $H$. 
We refer to the computed logits as ``known'' logits.

Step 3: approximate ``unknown'' logits for tokens in unselected clusters. 

Our head computes pseudo values for these logits \textit{without} loading their respective token heads. 
To do so, we utilize the probability invariant e.g., $\Sigma^k_{i=0} P_i = 1$
and the relationship between the sum of softmax probabilities and exponentials.
Given that known logits are $\text{logits}_{known}$ and unknown logits are $\text{logits}_{unknown}$, where the number of elements are $k$, and $N-k$, respectively, the derivation for $\text{logits}_{unknown}$ is as follows:
\begin{equation} 
    \begin{aligned} 
        P_{\text{known}} &= \frac{\text{logits}_{\text{known}}}{\text{logits}_{\text{known}} + \text{logits}_{\text{unknown}}}, \\ 
        \text{logits}_{\text{unknown}} &= \text{logits}_{\text{known}} \cdot \frac{1 - P_{\text{known}}}{P_{\text{known}}}. 
    \end{aligned}
\end{equation}

the first equation represents the softmax equation, and the second is its derived form. 
By calculating the derived form, our head determines the sum of unknown logits 
and accordingly assigns the mean value to each unknown logit.
Based on the union of known and unknown logits, our model samples a token for generation.

Such pseudo logits are vital to model perplexity: 
it ensures a smoother probability distribution over all tokens, regardless of whether their token clusters are selected or not.
By contrast, assigning arbitrary values (e.g., $-\infty$) results in very high perplexity at times.

\paragraph{Computational overhead considerations}
While pseudo logits does recover perplexity, their adds non-trivial wall clock delays.
(1) Irregularity of unloaded clusters. The unloaded clusters vary in size, which complicates the calculation of pseudo-logits.
(2) Scatter operations. After calculating pseudo-logits, the process involves scattering them back to the logit vector, which is computationally expensive.
Due to these overheads, we apply the hierarchical head judiciously, primarily on smaller models.

% !TeX root = main.tex

\section{Implementation}
\label{sec:impl}

\textbf{Training details}
We used the Pile dataset~\cite{pile} (200B) to train the baselines, our models, and hierarchical heads.
This dataset contains a diverse set of text data, including books, articles, and web pages.
It is the largest that we can afford (given our limited academic resource budget) to train the models.

We have trained our models by regularly submitting jobs to a SLURM cluster at the authors' institution. 
Each training job runs for at most 3 days (cluster policy) and on 4--6 A100 GPUs. 
As of this writing, we have trained ours models with the following amounts of tokens: 215B/196B for small/medium continual models, 200B/187B/92B for tiny/small/medium pretrained models.

\textbf{How is SVD trained}
For continual train, we inherited the official checkpoints and modify their layers by applying Eq~\ref{eq:continue-pretrain}.
For regular pretrain, we initialized models from scratch using random weights and biases, following Eq~\ref{eq:regular-pretrain}.
This training requires the end-to-end training; hence, the most time-consuming.

\textbf{How are sparsity predictors trained}
We have trained sparsity predictors for the channel-mix layers based on our pretrained models, which is a similar approach to the existing work~\cite{liu2023dejavu, xue2024powerinfer2fast}.
For training data, we recorded activations triggered by 5,000 input samples and accompanying weights $W_{k,v}$.
Unlike the SVD training, this requires training only two linear layers; thus, it is less time-consuming.
We have trained the predictors for at least 50 epochs for each model size, requiring approximately 24 GPU hours on an RTX 4090.

\textbf{How is head trained}
As mentioned in Sec~\ref{design:outer}, $H_1$ is the only trainable layer for hierarchical heads.
Each model was trained on approximately 1B tokens for at least 30 epochs.

\textbf{Custom ARM NEON kernels}
The official RWKV codebase is unoptimized for ARM CPUs.
To run INT8 inference on ARM CPUs, 
it falls back to Python's tensor conversions, resulting in more than a 10x slowdown.
To address this, we implemented custom NEON kernels that \textit{fuse dequantized and matrix-vector multiplications}. 
Our kernels support a variety of ARM devices, dequantizing to FP16 for chips supporting hardware NEON FP16 (e.g., RPi5), or dequantizing to FP32 for chips that do not support FP16 (e.g., RPi4 and earlier, Opi).
Our NEON kernels made the INT8 evaluation (Appendix) and our prototype (Figure~\ref{fig:demo}) possible. 
We will make them opensource.
% !TeX root = main.tex
\section{Experiments}
\label{sec:eval}
We conduct a comprehensive evaluation of \sys{} to assess its performance, focusing on the following aspects: (1) memory usage, (2) accuracy, and (3) inference speed.

\subsection{Methodology}

\paragraph{Benchmarks}
We evaluate on a wide range of language benchmarks; see Appendix~\ref{appx:all-results} for details. 
In this section, we present results from one of the most challenging benchmarks, \textit{lambada\_openai}~\cite{paperno2016lambadadataset}, which includes test cases requiring long-range contextual and semantic reasoning.
For other benchmarks, which are less challenging, \sys{} demonstrates even more significant advantages (see Appendix for details).

% !TeX root = main.tex

\begin{table}
    \centering
    \caption{Models used in the experiments. "Params" means the size of a checkpoint saved on disk, which differs from the memory usage. 
    }
    \includegraphics[width=0.38\textwidth]{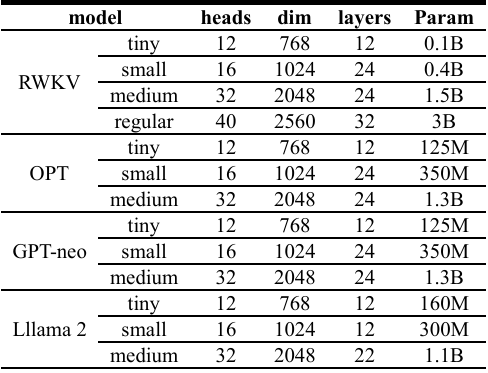} 
    \label{tab:model}
\end{table}

% !TeX root = main.tex

\begin{table}[t!]
    \centering
    \caption{CPU platforms for inference. opi2w is used in the prototype in Figure~\ref{fig:demo}.}
    \includegraphics[width=0.4\textwidth]{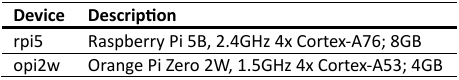}
    \label{tab:platform}
\end{table}

\paragraph{Model versions and baselines.}
We consider RWKV variants that are suitable for resource-constrained devices. 
Table~\ref{tab:model} lists these variants along with their hyperparameters.
Specifically, we compare the following implementations.
\begin{myitemize}
    \item \textbf{RWKV-vanilla}: the RWKV (v5) checkpoints released by the authors; unmodified. 
    \item \textbf{RWKV-ours}: we take \textit{RWKV-vanilla}, modify their architectures to ours, and perform continual training on the Pile dataset, updating all the model parameters.
    Projections are represented as simple SVD (\S\ref{sec:design:svd}). 
    \item \textbf{RWKV-ours-pretrain}: we initalize RWKV models with our architecture, where projections are represented as enhanvced SVD (\S\ref{sec:design:svd}), and pretrain them from scratch on the Pile dataset. The related results are in Appendix.
\end{myitemize}

We compare them to various transformer models of similar sizes and FLOPs requirements (listed in Table~\ref{tab:model}):
OPT~\cite{zhang2022optopena}, GPT-Neo~\cite{gpt-neo}, and smaller variants of Llama~\cite{miao2024specinferaccelerating,microllama, zhang2024tinyllama}.

\paragraph{Inference efficiency}
We execute inference with model weights in FP16 or INT8 and apply 0.7/0.8 for MLP and quantization-based predictors, respectively.
To measure the inference efficiency, we deploy models on two edge platforms, listed in Table~\ref{tab:platform}.

We characterize \textbf{the inference speed} as token-per-second (TPS) in text generation. 
We characterize a model's \textbf{memory footprint} under two popular loading strategies: 
(1) \textit{full loading}: as the inference code launches, it loads all the model parameters into the memory, 
and therefore avoids any disk IO at inference time; 
with our techniques, \textit{full loading} would load all the weights except those in embedding, FFNs (in channel-mix), and classification head, which are managed by our proposed techniques in Section~\ref{design:sparsity}--\ref{design:outer}. 
(2) \textit{layerwise loading}:
as a more aggressive approach, the inference code loads layer $N+1$ while executing layer $N$; this shrinks the memory footprint but nevertheless incurs high delays in disk IO at inference time.

% !TeX root = main.tex
\begin{figure}[t]
	\centering
	\includegraphics[width=0.35\textwidth]{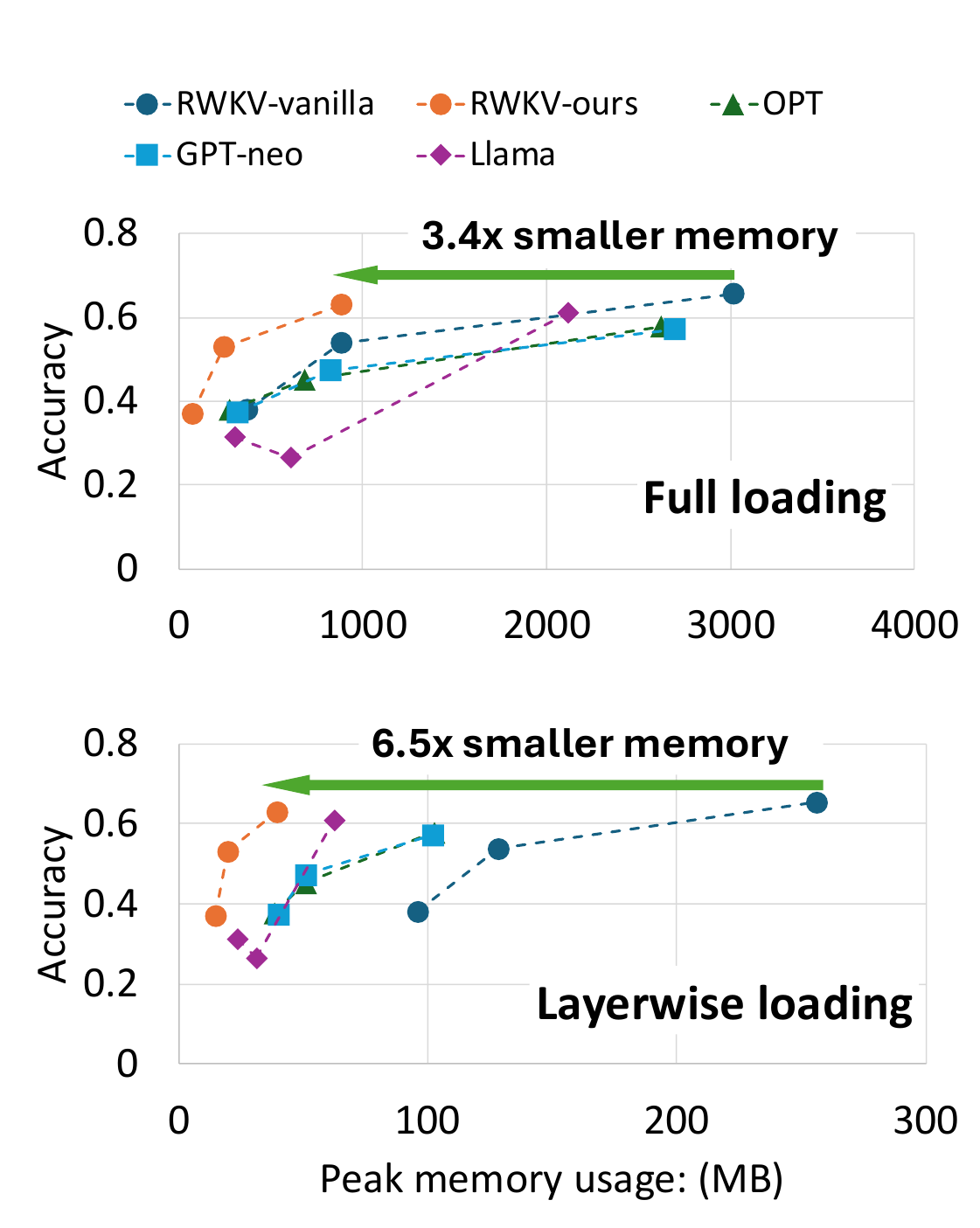}
	\caption{Accuracy \& memory footprint comparison between RWKV and transformer models.
        RWKV-ours has smaller memory footprint than other models and still maintain the comparable accuracy under both loading strategies. All model weights in FP16.
		Benchmark: lambada\_openai.
        }
	\label{fig:acc-rwkv}
	\vspace{-0.2in}
\end{figure}

\subsection{Model Memory footprint}
We consider a model's memory footprint as its maximum memory usage during model inference.
Results in Figure~\ref{fig:acc-rwkv} show that we significantly reduce memory footprint, while incurring 
little or no reduction in the inference accuracy that is defined by the benchmark tasks, 
e.g., accuracy defined in a word prediction task.
Our techniques also preserve model perplexity, which measures the model output fluency or coherence; see Appendix~\ref{appx:all-results} for details.

Figure~\ref{fig:acc-rwkv} shows 
that RWKV-ours \fcircle[white,fill=orange]{3pt}, 
compared to RWKV-vanilla \fcircle[white,fill=cerulean]{3pt}, 
reduces the memory for \textit{full loading} by 4x on average, 
and for \textit{layerwise loading} by 5x on average.
Meanwhile, RWKV-ours only experiences little accuracy drop, 
around 1pp, 0.9pp and 1.5pp for tiny, small and medium models, respectively.

We also tested a larger model, RWKV-regular (3B before our compression). 
At the time of writing (Jan '25) with only 100B tokens trained,  
we reduce the memory by 3.2x while only seeing accuracy 3pp lower in accuracy 
(vanilla is 0.68 whereas ours is 0.65).
With continual training, we expect this accuracy gap to diminsh.

\begin{figure}
	\centering
	\includegraphics[width=0.48\textwidth]{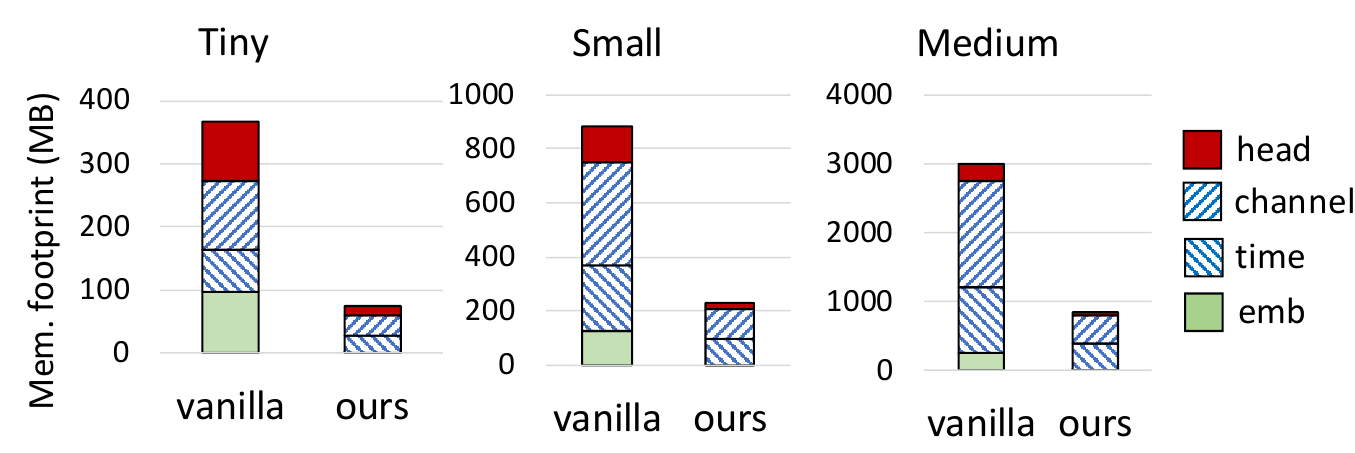}
	\caption{Memory breakdown of RWKV models, with the full loading strategy.
         Our models have significant reduction in all components e.g., embedding, time/channel-mix, and head.}
	\label{fig:breakdown-mem}
\end{figure}

Figure~\ref{fig:breakdown-mem} 
further breaks down our memory footprint by model components. 
Across all model sizes, our techniques (SVD and sparsity) significantly reduce the memory footprint of RWKV blocks, 
by 2.5x for the time-mix and 3.6x for the channel-mix.
In particular, for tiny and small models where the embedding and head layers are major memory consumers, 
our hierarchical heads reduce memory usage by 6.7x by loading only relevant weight clusters for predicted output tokens, 
and our embedding cache reduces memory by more than an order of magnitude by only loading embeddings for tokens in the context.
Without these two optimizations (and only with SVD and sparsity), we would not see much memory reduction for smaller models. 

\paragraph{Comparison to Transformer models}
Our RWKV models demonstrate a clear advantage against transformer models in terms of memory efficiency. 
Results are shown in Figure~\ref{fig:acc-rwkv}(a) and (b).
Note that this comparison \textit{favors} transformers by not counting 
their KV cache sizes; 
RWKV maintains compact, O(1) memory states across timesteps, and therefore does not require KV caches by design. 

At similar accuracy levels, RWKV-ours consume 3x less memory than transformer models, as exemplified by our medium model (the rightmost \fcircle[white,fill=orange]{3pt}) vs. Llama (the right most \fdiamond[white,fill=darkorchid]{3pt}) in the figures. 
With similar memory footprints, RWKV-ours achieves much higher accuracy than transformers, as shown by RWKV-ours vs. GPT-neo \fsquare[white, fill=electriccyan]{5pt} or Llama in the figures.
Much of the benefit comes from our techinques instead of RWKV itself: 
RWKV-vanilla, without our optimzations, would see lower or marginal memory savings, as shown by RWKV-vanilla vs. OPT \ftriangle[white, fill=forestgreen(web)]{5pt} or GPT-neo in the figures.

% !TeX root = main.tex

% !TeX root = main.tex

\begin{figure}
	\includegraphics[width=0.45\textwidth]{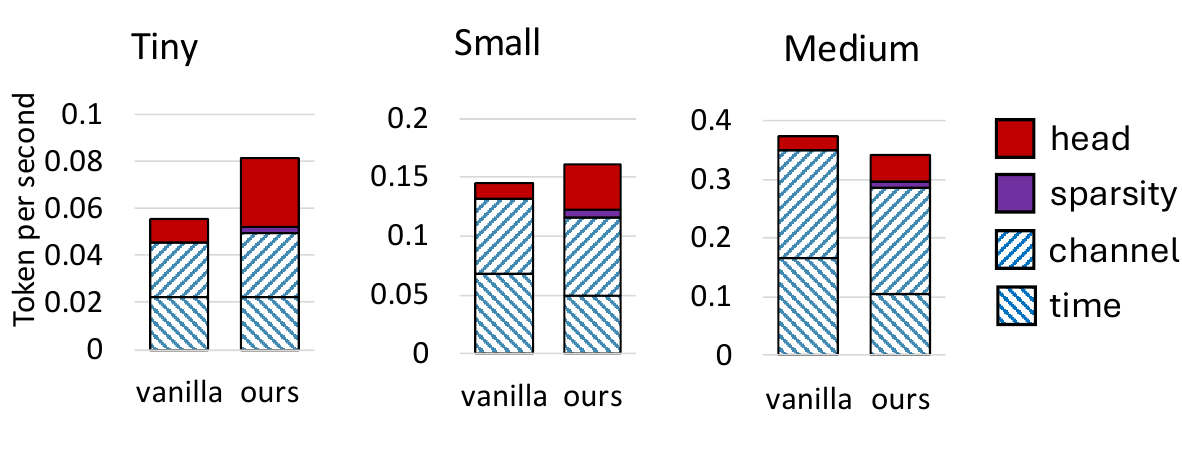}
    \caption{
        Inference time breakdown of RWKV-vanilla/ours, on the rpi5 hardware. 
        The head layer is the main difference between vanilla and ours;
        additionally, its layer becomes smaller on larger models while other layers remain the same.
    }
	\label{fig:breakdown-inference-time}
\end{figure}

\subsection{Inference speed}
Our memory optimization does not slow down inference much compared to RWKV-vanilla. 
As shown in Figure~\ref{fig:inference-time} (Appendix), 
Running FP16 small and medium models, RWKV-ours see a slight TPS drop or no degradation (5\% drop and 20\% increase for each model); 
on tiny model, the drop is more noticeable (29\%). 
The major factor of such slowdown, 
as illustrated in Figure~\ref{fig:breakdown-inference-time}, 
comes from hierarchical heads, which do gather-scatter operations on pseudo logits (lower parallelism, memory latency bound) in lieu of a full matrix multiplication (regular parallelism).
As model sizes increase, the overhead of hierarchical heads is dwarfed by those of RWKV blocks, 
diminishing the gap in inference speeds. 
% !TeX root = main.tex

\section{Related work}
\paragraph{Singular Value Decomposition (SVD)}
SVD has been extensively explored for compressing various layers~\cite{chen2018groupreduceblockwise, acharya2018onlineembedding,bennoach2020compressingpretrained}.
One recent work~\cite{hsu2022languagemodeld} reconstructs the decomposed transformer blocks applying the Fisher information.
However, none has explored its validity for the RWKV models.

\paragraph{Clustering}
Research on clustering ranges from unsupervised algorithm such as Kmeans to supervised~\cite{barnabo2023supervisedclustering} or semi-supervised approaches assisted by  LLMs~\cite{viswanathan2024largelanguage,tipirneni2024contextawareclustering}.
Such research targets on reducing clustering loss, which is orthogonal to \sys{}'s goal.
Unlike these cases, recent work~\cite{agarwalchaiclustered} leverages clustering to remove redundant attention heads on inference, which aligns with our idea; however our focus is on the output layer.
\paragraph{Quantization}
Recent work on LLM is the weight quantization by n-bits while minimizing errors in precision~\cite{frantar2023gptqaccurate,dettmers2023case4bit}.
Our design operates in the same domain and can be co-beneficial with the quantization.

\paragraph{Sparsity}
Active research on the transformer block is sparsity on Feed-Forward Networks, pruning unrelated their rows and columns~\cite{liu2023dejavu,xue2024powerinfer2fast,song2024turbosparsea}. 
The idea is proven to be effective in the transformer models; yet it is not explored in RWKV.

% !TeX root = main.tex
\section{Conclusions}
We present \sys{}, a suite of efficient compression techniques for RWKV, deployable on edge devices like Raspberry Pi 5. \sys{} reduces memory usage by 3.4x -- 5x while maintaining accuracy. Its computational overhead is negligible for edge deployment.

\bibliographystyle{icml2025}
% \bibliographystyle{plainnat}
% \bibliography{vldb_sample}  % vldb_sample.bib is the name of the Bibliography in this case
\bibliography{bib/rwkv, bib/rwkv-misc}

% appendix
\newpage
\appendix
\onecolumn
% !Tex root = main.tex

\section{Background}
\paragraph{Singular Value Decomposition (SVD)}
SVD is one of the low-rank factorizations to approximate an original weight matrix with standard factorization.
It decomposes the matrix into three matrices ($U, \Sigma, V$) and multiplies them to approximate the original one with fewer total parameters.

Given a matrix $W \in \mathbb{R}^{M \times N}$, SVD reconstructs it as follows:
\begin{equation}
\label{eq:svd}
W \approx U \Sigma V^T
\end{equation}
where $U \in \mathbb{R}^{M \times r}$, 
$V \in \mathbb{R}^{N \times r}$, and $r$ is a target rank for SVD. 
$\Sigma$ is a diagonal matrix of non-zero singular values $diag(\sigma_{1}, ..., \sigma_r)$, where $\sigma_1 \geq \sigma_2 \geq ... \geq \sigma_r \geq ... \geq \sigma_k > 0$, and $k$ is the rank of matrix $W$.
Setting zeros to $\sigma_{r+1}, ..., \sigma_{k}$ achieves the low-rank approximation.

%https://ocw.mit.edu/courses/18-06sc-linear-algebra-fall-2011/d273f75ee2552a5c3c35ccab37e5edce_MIT18_06SCF11_Ses3.5sum.pdf
To compute SVD of matrix $W$, we first determine $W^TW$ and find its eigenvalues and eigenvectors. 
The eigenvalues correspond to the squared singular values $\sigma^2$, so taking their square roots gives the singular values $\sigma$. 
The eigenvectors of $W^TW$ form the right singular matrix $V$
Next, we compute the left singular matrix $U$ using the relation $U=WV\Sigma^-1$, where $\Sigma$ is the diagonal matrix of singular values.
The final decomposition is given as $W \approx U\Sigma V^T$, where $U$ contains the left singular vectors, $\Sigma$ contains the singular values along its diagonal, and $V$ contains the right singular vectors.
% !Tex root = main.tex

\section{Additional evaluation}
\label{appx:all-results}
% !TeX root = main.tex
\begin{table}[h]
    \centering
    \caption{A collection of benchmark tasks for the model evaluation. Tasks are popularly used in the LLM evaluation.}
    \includegraphics[width=0.7\textwidth]{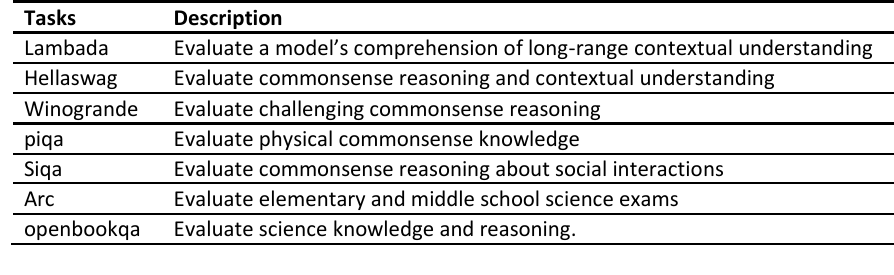} 
	\label{tab:benchmark}
\end{table}

% !TeX root = main.tex
\begin{table}[h]
    \centering
    \caption{Benchmark results for all models (acc = accuracy, ppl = perplexity).}
    \label{tab:all-results}
    \includegraphics[width=1\textwidth]{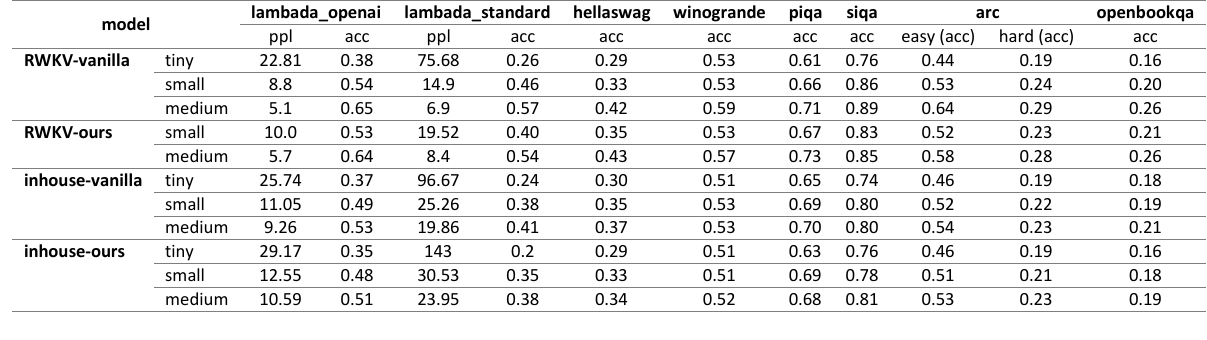} 
\end{table}

\subsection{Benchmark results for all models}
We report our full evaluations on the official and our models in Table~\ref{tab:all-results}.

\subsection{Energy consumption}
RWKV-ours consume slightly more energy per inference, compared to RWKV-vanilla.
As measured using a USB power meter, both variants, during active inference, draw the same device power (around 6.5 Watts for Rpi5).
Hence, the total energy consumption is proportional to the time taken to generate a certain number of tokens, 
with RWKV-ours consuming approximately 10\% more energy on the small models (e.g., 214J vs 195J for 200 tokens)

\subsection{Ablation study}
\label{eval:ablation}
We conduct ablation study to evaluate individual effectiveness of each technique.
The evaluation setting such as thresholds for sparsity predictors and hierarchical head is the same as the main evaluation in Section~\ref{sec:eval}.

\paragraph{Our optimizations impact on accuracy}
As shown in Table~\ref{tab:ablation-acc}, we observe that the ablated models show a slight drop in accuracy compared to RWKV-vanilla;
The losses are 1.3pp, 0.7pp and 2pp from tiny, small and medium models, respectively; the numbers are averaged across ablated models in the same parameter sizes.
Overall, among the three optimizations, SVD has the highest impact on model accuracy while Sparsity shows the least.
Notably, the accuracy of the SVD-ablated models are closely aligned with the vanilla models.
This is reasonable because ablating SVD essentially reverts the models to their vanilla counterparts.
Considering that the memory efficiency from HH diminishes as the model layers and dimensions scale up, we disable HH for medium or larger models. 
For a detailed breakdown of individual memory efficiency, we refer to the contents presented in Figure~\ref{fig:breakdown-mem}.
% !TeX root = main.tex
\begin{table}
	\centering
	\caption{Accuracy of ablated models. Benchmark: OpenAI lambada. 
	A component column refers to the model with all components except the specified component.
        Ablated models have slight drops in accuracy; yet, having more memory usage than full models (All).
    (SVD = Singular Value Decomposition, HH = hierarchical heads, Sparse = sparsity predictors) }
	\includegraphics[width=.4\textwidth]{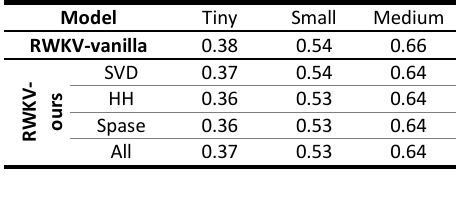}
	\label{tab:ablation-acc}
\end{table}

\subsection{Sensitive analysis}

% !TeX root = main.tex
\begin{table}[h]
	\centering
	\caption{Accuracy \& peak memory usage comparison of \textbf{inhouse} RWKV models.
        \textbf{inhouse-ours} has smaller memory footprint than \textbf{inhouse-vanilla} models and still maintain the comparable accuracy
        on both loading strategies. 
        }
	\label{fig:acc-inhouse}
	\includegraphics[width=0.48\textwidth]{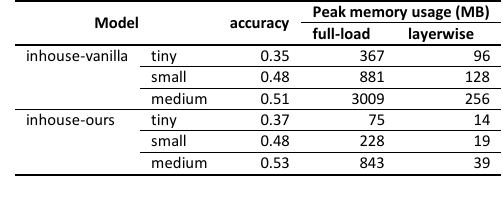}
\end{table}
% !TeX root = main.tex
\begin{figure}
    \centering
    \includegraphics[width=0.48\textwidth]{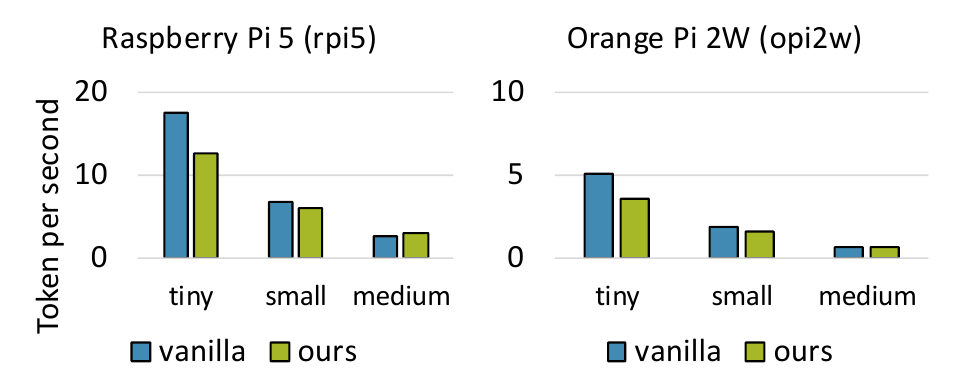}
    \caption{
	Comparison of TPS between \textbf{inhouse-vanilla} and \textbf{inhouse-ours} on rpi5 and opi2w. 
        \textbf{inhouse-ours} exhibits a slight loss, compared to \textbf{inhouse-vanilla}. This loss becomes minor in the small model. 
    }
    \label{fig:infer-inhouse}
\end{figure}

\paragraph{SVD as a suitable architecture for pretraining.}
We test the idea, Eq~\ref{eq:regular-pretrain}, in section~\ref{sec:svd}: replacing $W \in R^{D \times D}$ projection matrices with their SVD decomposition format, 
which is further enhanced with non-linearity, composed by full-rank, diagonal matrices. 
We initialize such a model architecture from scratch and pretrain them with the Pile datasets. 
We named these models, pretrained by us from scratch, as ``in-house'' checkpoints:
\textbf{inhouse-vanilla}: models in the vanilla architecture without our optimiziations. 
\textbf{inhouse-ours}: model architectures with our optimizations.

As Table~\ref{fig:acc-inhouse} shows, 
the comparison between \textbf{inhouse-vanilla} and \textbf{inhouse-ours} manifests that the SVD can be a suitable choice. 
While reducing the parameters of project matrices by 4x (and the total model sizes by 3.5x--4.8x, ranging from tiny to medium models),
the total accuracy sees, rather, slight gains: 1.4pp on average.
The additional FLOPS required by SVD (at inference time) is also negligible: 
as shown in Figure~\ref{fig:infer-inhouse}, \textbf{inhouse-ours} is only 13.7\% slower than \textbf{inhouse-vanilla} on average on rpi5, and 20\% slower on opi2w. 
Furthermore, we observed little to none training throughput difference on these two \textbf{inhouse-vanilla} and \textbf{inhouse-ours}.
This suggest that SVD should be considered as a ``free'' size optimization for pretraining RWKV models. 

We notice an important caveat, though. 
 \textbf{inhouse-vanilla}, with our trainig budget and datasets, fall under the accuracy of the official checkpoints 
 (for instance, on \textit{lambada\_openai}, \textbf{inhouse-vanilla} show lower accuracy by 7.7pp from tiny to medium models. 
We attribute the reason as the official checkpoints were trained for far more tokens (1.12 T, 5x more than ours) 
and datasets (RWKV authors disclosed their choices of their training data, but not the exact ratios or scripts~\cite{peng2023rwkvreinventing}). 

\paragraph{The choice of low-rank approximation factors}
We tested aggressive (16x) and light (4x) SVD decomposition factors to find the optimal balance between memory efficiency and accuracy. 
We find that the 16x factor results in detrimental accuracy; on the contrary, the light decomposition (4x) brings a slight or no accuracy improvement, compared to 8x default decomposition.
In detail, 16x shows significant drops: 2.85pp for the tiny model, 11pp for the small model, and 29pp for the medium model; 4x models achieve very similar accuracy to 8x with less than 1pp.
For the light decomposition, albeit comparable accuracy, it still provides complementary benefits with quantization regarding to memory efficiency.

% !TeX root = main.tex

\begin{figure}
    \centering
    \includegraphics[page=1,width=0.48\textwidth]{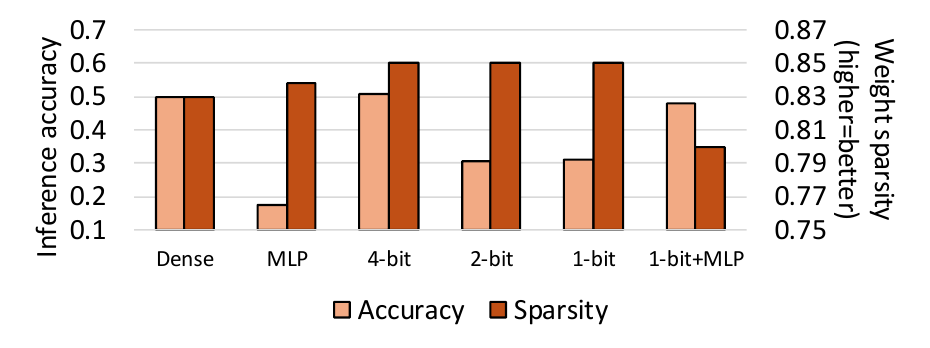}
    \caption{Accuracy and sparsity rate achieved by the ground-truth (GT) and quantized predictors ($n$-bit). Benchmark: lambada\_openai}
    \label{fig:results-sparsity-predictor}
\end{figure}
\paragraph{A variety of sparsity predictors}
We find that a deeply quantized predictor (1-bit), when ensembled with a learning-based MLP, can provide the best of both worlds: small predictor size and high neuron recall. This combination outperforms using either the MLP or a larger quantized predictor (e.g., 4-bit) alone. 

To show this, we study a range of sparsity predictors and their ensembles on our small model by executing inference for the OpenAI benchmark.
As illustrated in Figure~\ref{fig:results-sparsity-predictor},
n-bit quantized networks, which constitute 1/n of the original FP16 FFN size, can predict the sparsity rate and accuracy close to the ground truth (83\% vs 85\%), while aggressive quantized networks (1/2-bits) lose their accuracy by a half of that of 4-bit network despite their similar sparsity rate.
However, we found that these accuracy losses from heavy quantization can be mitigated by ensembling the quantized predictor and MLP.
Specifically, one such ensemble (1-bit and MLP) leads to losing minor accuracy degradation (1.7pp) and having less memory usage due to aggressive quantization, compared to the sole 4-bit quantized network.
This indicates that the 4-bit predictor w/o MLP is more accurate, while the 1-bit predictor w/ MLP is more memory efficient, which suggests that the choice of predictor depends on the user's priorities, whether favoring memory reduction or higher accuracy, and highlights the trade-off between these two factors.

\paragraph{The number of clusters}
We find that applying appropriate thresholds for hierarchical heads is crucial to achieve the best accuracy and memory efficiency.
To find out its implication, we varied its threshold to load more (0.99) or less (0.85) number of clusters; our default is 0.95, which is empricially determined at best.
We observe that 0.85 reduces the memory usage by 2x, yet dropping 10pp in accuracy.
Conversely, increasing the value to 0.99 loads 2x more clusters, which volumes the memory usage up by 2x; yet, this improves 1.5pp in accuracy.
Note that the described memory usage is derived from only hierarchical head size, not from the entire model.
Our results implicate that the number of clusters should be carefully determined to balance the trade-off between the memory usage and the accuracy.
% !TeX root = main.tex
\begin{figure}[h]
	\centering
	\includegraphics[width=0.48\textwidth]{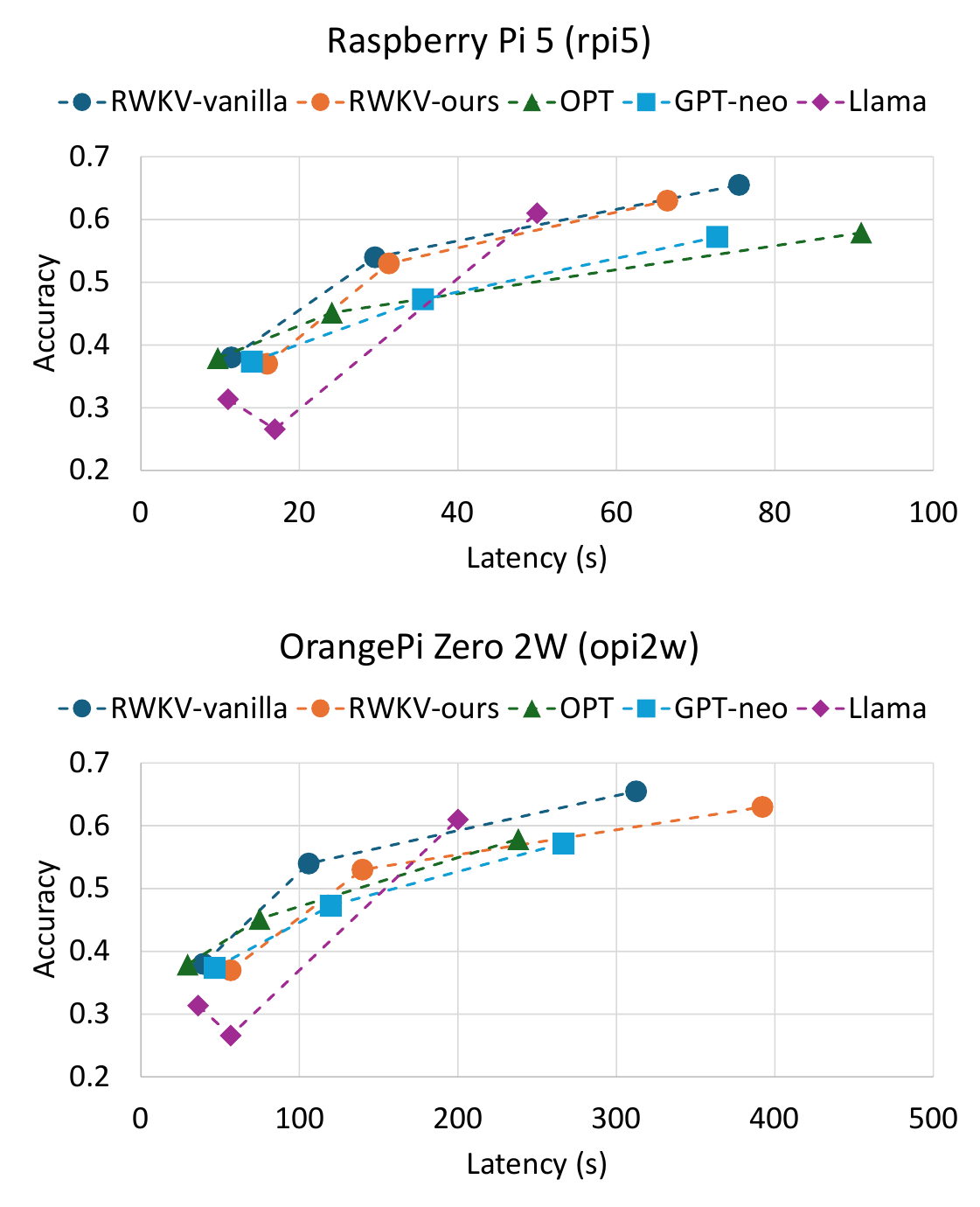}
	\caption{Model comparison between transformer and RWKV models on both CPU platforms.
        The model sizes are tiny, small, and medium from left to right.
        RWKV-ours are the optimal models among displayed models considering all crucial metrics e.g., accuracy, peak memory usage, and TPS.}
	\label{fig:tf-mem-tps}
\end{figure}

\subsection{Inference: Transformer vs. RWKV.}
RWKV-ours have slight drops or no degradation in token-per-second (TPS) at similar accuracy levels, such as for medium and small models.
While RWKV-vanilla has higher accuracy than any other models, its benefits in memory reduction and TPS are minor.
As demonstrated in Figure~\ref{fig:tf-mem-tps}, RWKV-ours results in 19\% drops and 7\% gains in TPS on average, compared to other small and medium transformer models on the rpi5; by contrast, executing our tiny model drops 28\%.
This is understandable because our models require more GFLOPs due to additional multiplications induced by our augmented layers.
While RWKV-ours have little benefits on TPS, our models are more optimal other transformer models with a similar or higher accuracy and huge amount of memory savings.
On the opi2w, we found that tendencies of the resultant numbers are very homogeneous to those of rpi5; therefore, we omit its description.
% !TeX root = main.tex

\begin{figure}[h]
	\centering
	\includegraphics[width=0.48\textwidth]{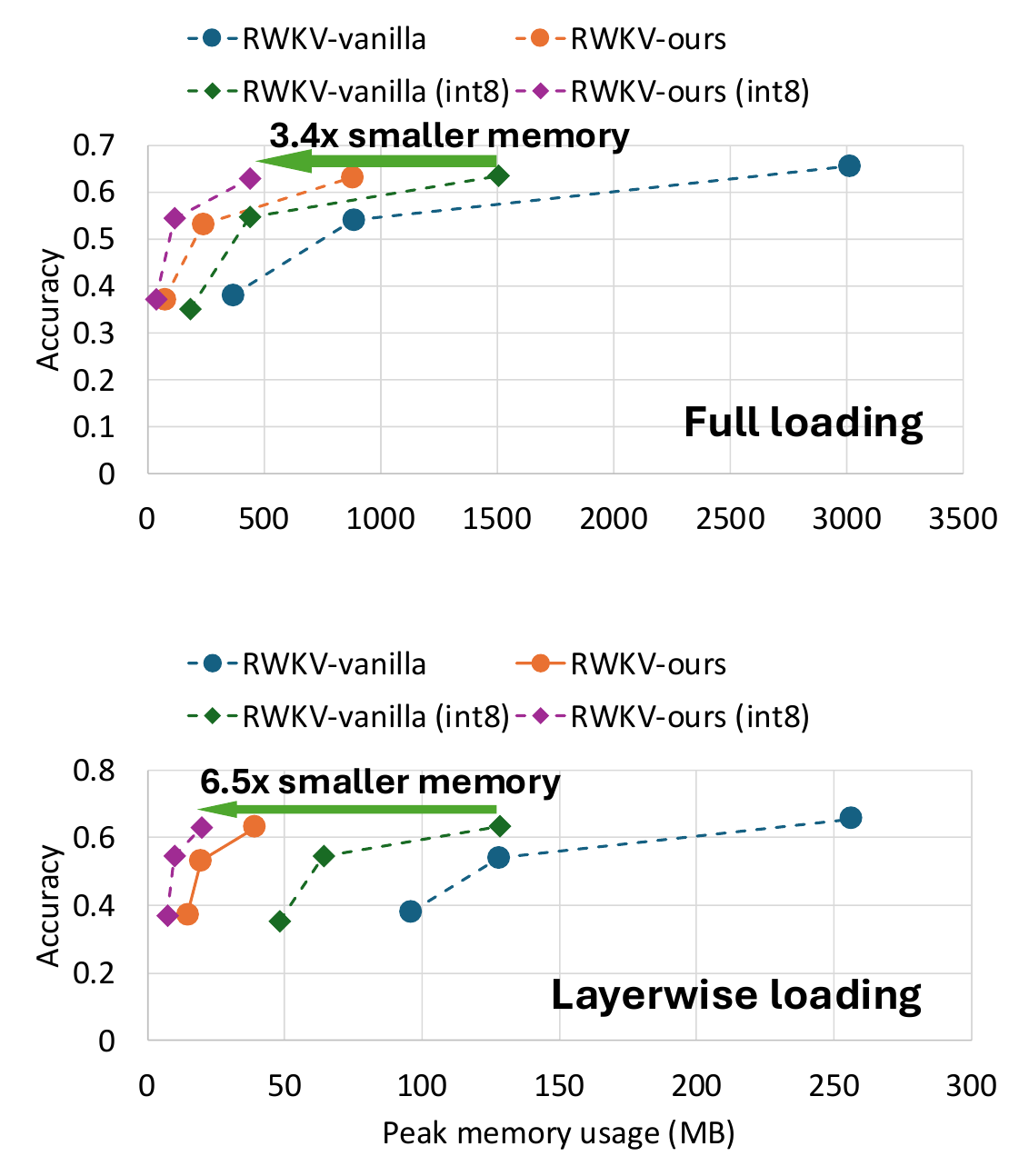}
	\caption{Comparison of accuracy and memory usage between float16 and int8 for RWKV models. 
        Our models are complement to quantization without a significant performance drop.
		}
	\label{fig:acc-i8}
\end{figure}

\subsection{Ours vs. INT8 quantization}
\paragraph{Compatibility with quantization (INT8)}
Our optimizations complement model quantization effectively. 
Combined with a popular quantization scheme INT8, 
this results in a 10x reduction in memory footprint on average across different parameter sizes (of which 2x from quantization and around 5x from our optimizations).

To demonstrate this, we compare RWKV-vanilla and RWKV-ours, both before and after quantization.
As shown in Figure~\ref{fig:acc-i8}, RWKV-ours benefits from INT8 quantization by reducing memory usage by roughly 2x while incurring minimal in accuracy (less than 1pp on both small and medium models).
This indicates that RWKV-ours is as robust, if not more so, to quantization compared to RWKV-vanilla, which sees an average loss of 1.5pp across different parameter sizes.

% !TeX root = main.tex

\begin{figure}
    \centering
	\includegraphics[width=0.5\textwidth]{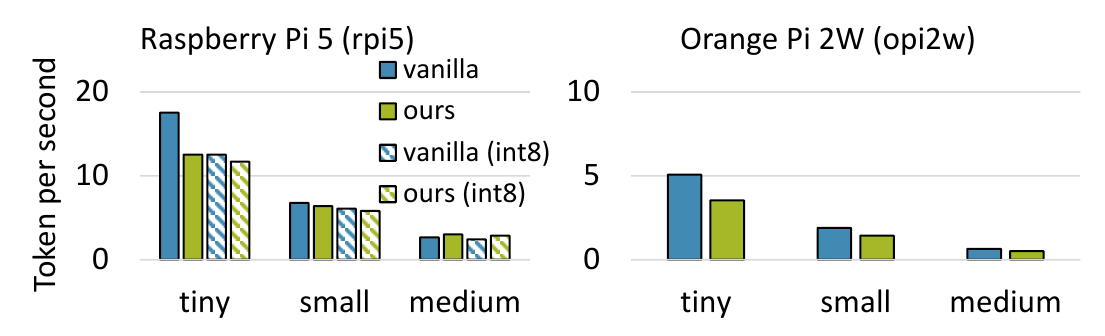}
    \caption{
	    Comparison of TPS between RWKV-vanilla and RWKV-ours on two devices. 
        RWKV-ours exhibits a slight loss, compared to RWKV-vanilla. This loss becomes minor in the small model. 
    }
	\label{fig:inference-time}
\end{figure}

\paragraph{Inference speed.}
For INT8 inference, RWKV-ours and vanilla show similar inference speeds to the FP16 inference with minor TPS drops.
Note that this is a remarkable achievement thanks to our NEON kernels; Turning off the kernels leads a detrimental effect on the speed (10x slower).
As shown in Figure~\ref{fig:inference-time}, all sizes from tiny to medium RWKV-ours show a minor decrease in TPS, which is 7\%, 9\%, and 5\%, respectively. 
RWKV-vanillas show similar performance drops (10\% and 9\%) on small and medium models; on the contrary, running the tiny vanilla results in a drop in 40\% TPS.
These performance drops in INT8 compared to FP16 are due to under-optimization, e.g., cache alignment for a specific instruction.
We will plan to close this gap in future work.


% That's all folks!
\end{document}